\documentclass[12pt]{article} 
 
\usepackage{amsmath,amssymb} 
\usepackage{vmargin} 
\usepackage[dvips]{graphicx} 
\usepackage{psfrag} 
\usepackage{notations1} 
\usepackage[bold]{bbdd} 
\usepackage{times} 
\usepackage{setspace} 
\usepackage{pifont} 
\usepackage{array} 
\usepackage{mathptmx} 
\usepackage{subfigure}

\newcommand\vek[1]{\mathbf{#1}}

 \setpapersize{A4} 
 \setmarginsrb{3cm}{2cm}{3cm}{2cm}{1cm}{0cm}{0cm}{1.5cm}

\begin{document} 
 
\begin{center}
{\Large Novel anisotropic continuum-discrete damage model capable }\\
\vspace{1mm}
{\Large of representing localized failure of massive structures. Part II:}\\
\vspace{1mm}
{\Large Identification from tests under heterogeneous stress field }\\
\vspace{6mm}

A.\,Ku\v{c}erov\'a$^1$, D.\,Brancherie$^2$, A.\,Ibrahimbegovic$^3$, J.\,Zeman$^1$ and Z.\,Bittnar$^1$
\\
\vspace*{6mm} 
$^1$
Department of Structural Mechanics, Faculty of Civil 
Engineering, Czech Technical University in Prague,
Th{\' a}kurova 7, Prague, Czech Republic
\\
\vspace*{1mm} 
$^2$
Laboratoire Roberval, Universit\'e de Technologie de Compi\`egne, \\
BP 20529, 60205 Compi\`egne Cedex, France
\\
\vspace*{1mm} 
$^3$
Ecole Normale Sup\'erieure de Cachan, 
Laboratoire de M\'ecanique  et Technologie, \\
61, av. du pr\'esident Wilson 94235 Cachan, France
\end{center}
\vspace*{5mm}
  
\begin{abstract} 
In Part I of this paper we have presented a simple model capable 
of describing the localized failure of a massive structure. In 
this part, we discuss the identification of the model parameters 
from two kinds of experiments: a uniaxial tensile test and a 
three-point bending test. The former is used only for illustration 
of material parameter response dependence, and we focus mostly 
upon the latter, 
 discussing the inverse optimization problem for which the specimen is 
subjected to a heterogeneous stress field. 
\end{abstract}

\section{Introduction} 
 
Present work is the follow-up on Part I of this paper 
\cite{Brancherie:2007}, discussing a relatively simple model 
capable of describing the behavior of a massive structure until 
the point of~localized failure. The model contains all the 
ingredients for taking into account both the~diffuse damage 
mechanism, which leads to the appearance of microcracks, as well 
as~the~failure process characterized by the propagation of 
macrocracks. Perhaps the most important advantage of the proposed 
model is the fact that all its parameters have a clear physical 
interpretation and can be straightforwardly visualized in terms of 
the shape of a stress-strain diagram. In addition, influence of 
each parameter is dominant only for a specific, easily 
recognizable, stages of material behavior. This kind of a priori 
knowledge has a potential to greatly simplify the model 
calibration and will be systematically used throughout the~paper. 
 
In Part II of this paper, we discuss the identification of the 
model parameters from experimental measurements made on a 
structural level. Generally speaking, the complexity of the 
identification procedure is determined by the choice of 
experimental setup. Solely from the identification point of view, 
the simplest experiment to execute is the uniaxial tensile test. 
In this case, the strain field stays mostly homogeneous during the 
whole procedure and the global response represented by the 
load-displacement diagram is very similar to the stress-strain 
curve for one material point; see Section \ref{tensile} for more 
details. The~model parameters can be then directly determined from 
the shape of the load-displacement curve. Such a uniform loading 
is, however, very difficult if not impossible to impose in a 
laboratory test, especially for quasi-brittle materials. 
Therefore, other tests are often used in~experimental practice. 
 
The three-point bending test, in particular, is considered to be much 
simpler to perform and its results are well-reproducible. Therefore, 
we focus on the identification procedure for the proposed model 
parameters directly from results of three-point bending test. Main 
difficulty is in this case imposed by heterogeneity of the stress and 
the strain fields, which is present since the very start of the 
experiment. The macro-scale measurements provide the load-deflection 
curve that integrates data from different parts of the specimen 
experiencing different regimes of (in)elastic behavior. For that 
reason, the possibility of a simple determination of model parameters 
from load-deflection curve is lost and an advanced calibration 
procedure needs to be applied. 
 
To take advantage of the model specific structure, already mentioned 
above, the identification procedure should be divided into three 
sequential stages discussed in detail in~Section~\ref{tpb}. From the 
algorithmic point of view, the material calibration can then be 
understood as~a~sequential optimization problem. Such 
approach has two main advantages: first, solving three simpler 
identification steps in a batch form is typically much more efficient 
then the full-scale problem; second, it allows to use only a subset of 
simulations for initial stages of identification process. 
 
A variety of techniques is available to identify material parameters
via optimization methods, see e.g.~\cite[and reference
therein]{Mahnken:2004:ECM}.  The gradient-based methods are usually
considered to be the most computationally efficient optimization
algorithms available and~as~such have been successfully used in a
variety of identification problems, e.g.
\cite{Mahnken:1996:PIVP,Iacono:2006:EMP,maier:2006:ijf}. 
For the~current model, however, analytic determination of sensitivities 
is fairly difficult, mainly due the history dependency of the model as
well as complex interaction of individual parameters. The accuracy of
numerical approximation to the 'exact' sensitivities, on~the~other
hand, is driven by the choice of pseudo-time step used in numerical
simulations. Clearly, to reduce the computational time, the
pseudo-time step should be used as large as possible. Therefore, the
response-based objective function will not be smooth and
gradient-based methods are unlikely to be very successful.
 
As an alternative, techniques of soft-computing can be employed 
for optimization of complex objective functions. For example, 
stochastic evolutionary algorithms have been successfully used for 
solution of identification problems on a level of material point 
\cite{Furukawa:1997:ijmne,Pyrz:2007:msmse} or~on~a~level of simple 
structures \cite{Ibrahimbegovic:2004,Leps:2005:eaiteo}. For the 
current case, however, complexity of~the~optimization can be 
attributed rather to its non-smooth character than to the 
appearance of multiple optima; the family of problems where 
evolutionary algorithms are the most successful methods. This 
opens the way to more specialized tools, which deliver higher 
efficiency when compared to usually time-consuming evolutionary 
algorithms. 
 
The approach adopted in the present work is based on an adaptive 
smoothing of the objective function by artificial neural 
networks~(see, e.g.,~\cite{Pichler:2003:BAMP,Waszczyszyn:2006:NIA} 
for alternative ANN-based solutions to identification problems). 
In particular, the approximated model is provided by the Radial 
Basis Function Network, described in Section~\ref{rbfn}, 
dynamically evolved by~minima located by a real-encoded genetic 
algorithm, briefly reviewed in Section~\ref{GRADE}. The proposed 
sequential numerical strategy is systematically verified in 
Section~\ref{results} with~attention paid to a detailed assessment 
of the proposed stochastic algorithm reliability. Final remarks 
and conclusions can be found in Section \ref{conclu}.

\section{A brief description of the identified model} 
 
In the present section, we give a brief description of~the model on 
which the identification procedure is based. For the readers interested in 
more details, the complete description of~the~model is given in Part I 
of this paper. 
 
As already mentioned, the proposed model is capable of taking into 
account two different types of dissipation (e.g. see 
\cite{adnan06}): 
\begin{itemize} 
  \item 
a bulk dissipation induced by the appearance of 
  uniformly distributed microcracks. This bulk dissipation is taken 
  into account by the use of a classical continuum damage model; 
\item 
  a surface dissipation induced by the development of macrocracks 
  responsible for~the~collapse of the structure. As presented in Part 
  I of this paper, this phase is taken into account by the use of a 
  strong discontinuity model. The surface dissipation is taken into 
  account by the introduction of a 
  traction/displacement jump relation. 
\end{itemize} 
 
Therefore, two different models are involved in the constitutive description: 
the one associated with the bulk material and the one associated with 
the displacement discontinuity. Both are built on the same scheme 
considering the thermodynamics of continuous media and interfaces. 
 
The key points of the construction of each of the two models are summarized in Table 
\ref{tab:tab_continuum} and Table \ref{tab:tab_discrete}. 
 
\begin{table}[htbp] 
\begin{center} 
\begin{tabular}{l|c} 
    \hline 
    Helmholtz energy    & $\bpsi(\bldepsbar,\bldDb,\bxi)=\frac{1}{2} 
    \bldepsbar : \bldDb^{-1} :\bldepsbar + \bar{\Xi}(\bxi)$ \\ 
  \hline 
    Damage function & $ \bphi(\bldsig,\bq)= \underbrace{\sqrt{\bldsig : \bldDe : 
    \bldsig}}_{\normde{\bldsig}} 
    - \frac{1}{\sqrt{E}}(\sig_f-\bq) $ \\ 
  \hline 
    State equations & $\bldsig = \bldDb^{-1} : \bldepsbar$ ~~ and ~~ $\bq= - 
    \frac{d}{d \bxi}\bXi(\bxi)$ \\ 
     \hline 
    Evolution equations &  $\dot{\bldDb} = \dot{\bgam} 
    \frac{\partial \bphi}{\partial \bldsig} \otimes \frac{\partial 
    \bphi}{\partial \bldsig} \frac{1}{\normde{\bldsig}}$ ~~;~~ $\bxi = 
    \dot{\bgam} \frac{\partial \bphi}{\partial \bq}$ \\ 
    \hline 
    Dissipation & $0 \leq \bar{\mathcal{D}}= \frac{1}{2} \dot{\bxi} (\bar{\sig}_f - \bK \bxi) $ \\ 
    \hline 
\end{tabular} 
\end{center} 
\caption{Main ingredients of the continuum damage model} 
\label{tab:tab_continuum} 
\end{table} 
 
\begin{table}[!ht] 
\begin{center} 
\begin{tabular}{l|c} 
    \hline 
    Helmholtz energy  & 
    $ \bbpsi(\bldjump,\bldQbb,\bbxi)=\frac{1}{2} \bldjump \cdot \bldQbb^{-1} \cdot 
  \bldjump + \bbXi(\bbxi) $\\ 
  \hline 
    Damage functions & $\bbphi_1(\bldt,\bbq) = \bldt \cdot 
    \bldn - (\bar{\bar{\sig}}_f - \bbq) $ \\ 
    & $\bbphi_2(\bldt,\bbq)= \vert \bldt \cdot \bldm \vert - (\bar{\bar{\sig}}_s - 
  \frac{\bar{\bar{\sig}}_s}{\bar{\bar{\sig}}_f}\bbq) $ \\ 
  \hline 
    State equations & $\bldt = \bldQbb^{-1} \cdot \bldjump$ ~~ and 
    ~~$ 
      \bbq = - \frac{\partial \bbXi}{\partial \bbxi}$\\ 
     \hline 
   Evolution equations & $\dot{\bbQ} = \dot{\bbgam}_1 \frac{1}{\bldt\cdot\bldn} + \dot{\bbgam}_2 \frac{1}{\vert\bldt \cdot\bldm\vert}  $ ~~;~~ $\dot{\bbxi} = \dot{\bbgam}_1 + \frac{\bar{\bar{\sig}}_s}{\bar{\bar{\sig}}_f} \dot{\bbgam}_2$ \\ 
    \hline 
    Dissipation & $0 \leq \bar{\bar{\mathcal{D}}} = \frac{1}{2} \dot{\bbxi} (\bar{\bar{\sig}}_f - \bbK \bbxi) $ \\ 
    \hline 
\end{tabular} 
\end{center} 
\caption{Main ingredients of the discrete damage model} 
\label{tab:tab_discrete} 
\end{table} 
 
For the discrete damage model, the isotropic softening law is 
chosen as: 
\begin{equation} 
  \bbq = \bar{\bar{\sig}}_f \left[ 1 - exp\left( - 
  \frac{\bar{\bar{\beta}}}{\bar{\bar{\sig}}_f} \bbxi\right)\right] 
\end{equation} 
 
In Tables \ref{tab:tab_continuum} and \ref{tab:tab_discrete} the 
variables $\dot{\bgam}$, 
$\dot{\bbgam}_1$ and $\dot{\bbgam}_2$ denote Lagrange multipliers induced by~the~use 
of the maximum dissipation principle. Moreover, $\bldjump$ denotes the 
displacement jump on the~surface of discontinuity. Finally, $\bldDb$ and $\bldQbb$ 
correspond to the damaged compliance of~the~continuum and discrete model, respectively.

Note that in a three-point bending or a uniaxial tensile test, the 
simulated response is almost independent of the limit tangential 
traction $\bar{\bar{\sig}}_s$. Therefore, its value was set to 
0.1$\bar{\bar{\sig}}_f$. With such simplification, there are six 
independent material parameters to be identified: 
\begin{itemize} 
  \item the elastic parameters: the Young modulus $E$ and the Poisson ratio $\nu$; 
  \item the continuum damage parameters : the limit stress $\bar{\sig}_f$ and 
  the hardening parameter $\bK$; 
  \item the discrete damage parameters : the limit 
  normal traction $\bar{\bar{\sig}}_f$,  and the softening parameter $\bar{\bar{\beta}}$. 
\end{itemize} 
 
The limits of realistic values for each parameter are shown in the 
Table \ref{tab_bound}. Note that in~our identification methodology we 
do not suppose to have an expert capable of giving the~initial 
estimate of material parameters values, as in 
e.g. \cite{Iacono:2006:EMP,Novak:2006:ANNI,maier:2006:ijf}. Therefore, 
the bounds on~model parameters were kept rather wide.

\begin{table} 
\center 
\begin{tabular}{rcl} 
$E$                  & $\in$ & (25.0, 50.0) GPa \\ 
$\nu$                & $\in$ & (0.1, 0.4) \\ 
$\bar{\sig}_f$       & $\in$ & (1.0, 5.0) MPa \\ 
$\bK$                & $\in$ & (10.0, 10000.0) MPa \\ 
$\bar{\bar{\sig}}_f$ & $\in$ & ($\bar{\sig}_f$+0.1, 2$\bar{\sig}_f$) MPa \\ 
$\bar{\bar{\beta}}$  & $\in$ & (0.1$\bar{\bar{\sig}}_f$, 10.0$\bar{\bar{\sig}}_f$) MPa/mm\\ 
\end{tabular} 
\caption{Limits for the model parameters.} 
\label{tab_bound} 
\end{table}

\section{Tensile test}\label{tensile} 
 
\begin{figure}[h]
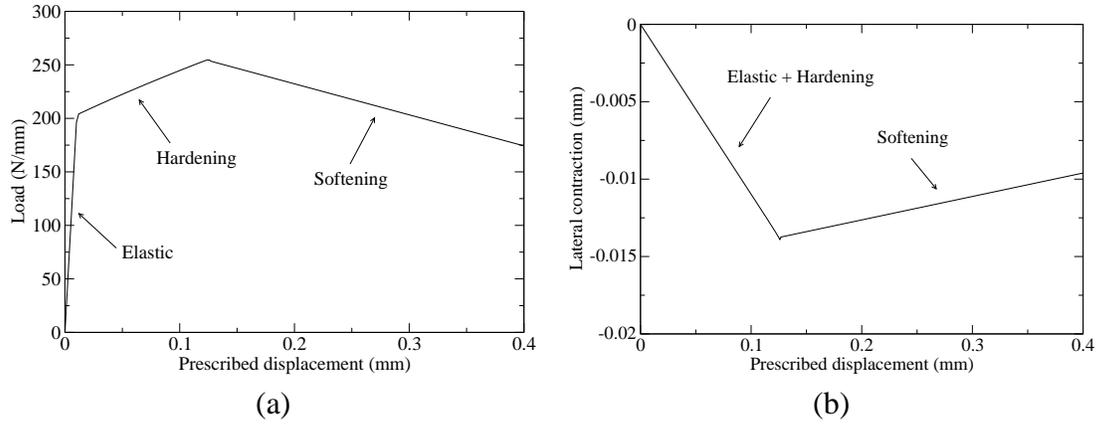
 
\center 
\begin{tabular}{cc} 
\includegraphics*[width=7cm,keepaspectratio]{tensile/fu.eps} & 
\includegraphics*[width=7cm,keepaspectratio]{tensile/dv.eps} \\ 
(a) & (b) 
\end{tabular} 
\caption{Tensile loading test: (a) Load-deflection diagram (b) Evolution of lateral contraction.} 
\label{fig_tensile} 
\end{figure}

The simplest possibility to identify material parameters for a 
particular concrete is to~perform a uniaxial tensile test. In this 
case, the stress and strain fields within the specimen would 
remain homogeneous until the final localized failure phase, and 
the behavior on~the~structural level is very close to the response 
of a material point. The load-displacement diagram consist of 
three easily recognizable parts, as shown in Figure 
\ref{fig_tensile}a: the first one corresponding to the elastic 
response of the material, the second one describing the hardening 
and the third part the softening regime. The calibration of model 
parameters can be carried out to follow the same pattern: first, 
Young's modulus $E$ and~Poisson's ratio $\nu$ are determined from 
the elastic part of the load-displacement diagram, followed 
by~the~limit stress $\bar{\sig}_f$ and the hardening parameter $\bK$ 
identification from the~part of~the~diagram corresponding to the 
hardening regime and, finally, the limit normal traction 
$\bar{\bar{\sig}}_f$ and~the~softening parameter 
$\bar{\bar{\beta}}$ are estimated from the softening branch. Note 
that for~ Poisson's ratio identification, one additional local 
measurement is needed to complement the~structural 
load-displacement curve, namely the measurement of lateral 
contraction of~the~specimen, see Figure \ref{fig_tensile}b. 
 
Although this kind of calibration procedure is robust and accurate 
\cite{Kucerova:2006:topping}, the experiment dealing with a simple 
tension test is rather difficult, if not impossible, to perform 
in~a~well-reproducible way. For that reason, we turn to study the 
possibility of parameter estimates by using three-point bending 
tests, which is much simpler to practically perform in~a~laboratory.

\section{Three-point bending test}\label{tpb} 
\begin{figure}[h]
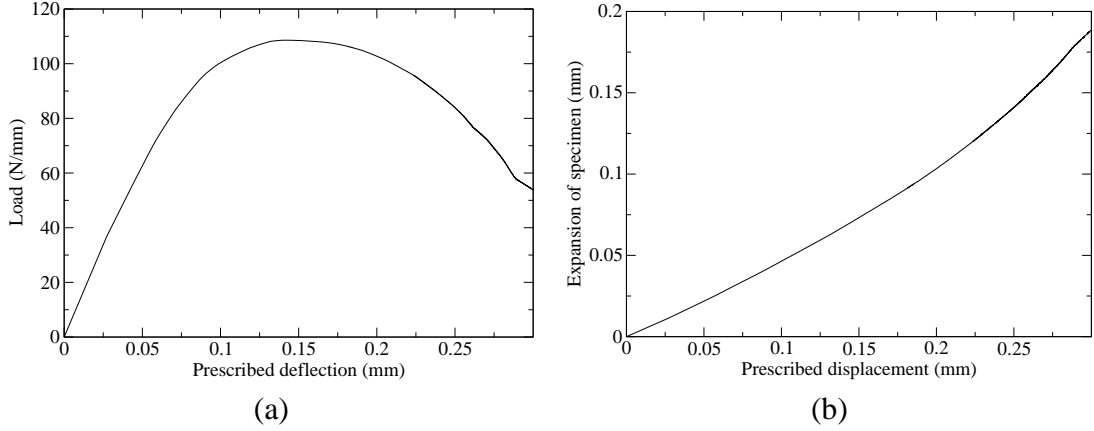
 
\center 
\begin{tabular}{cc} 
\includegraphics*[width=7cm,keepaspectratio]{bend/fu.eps} & 
\includegraphics*[width=7cm,keepaspectratio]{bend/du.eps} \\ 
(a) & (b) 
\end{tabular} 
\caption{Three-point bending test: (a) Load-deflection diagram (b) Evolution of expansion of specimen.} 
\label{fig_bend-ref} 
\end{figure}

In the case of a three-point bending test the global response of a 
specimen represented by~the~load-deflection ($L$-$u$) diagram for 
the structure cannot be simply related to three-stage material 
response with elastic, hardening and softening part (see Figure 
\ref{fig_bend-ref}a). Nevertheless, we assume that it will be 
still possible to employ the three-stage approach developed for 
the uniaxial tensile experiment. Similarly to the previous case, 
the solution process will be divided into the optimization of 
elastic, hardening and softening parameters in the sequential way. 
Each step is described in detail in the rest of this section. 
 
Due to lack of experimental data, a reference simulation with 
parameters shown in Table~\ref{tab_optimum} will be used to 
provide the target data. These are the same values as considered 
for~simulation presented in Part I of this paper 
\cite{Brancherie:2007}. 
 
\begin{table} 
\center 
\begin{tabular}{rcl} 
$E$                  & = & 38.0 GPa \\ 
$\nu$                & = & 0.1 \\ 
$\bar{\sig}_f$       & = & 2.2 MPa \\ 
$\bK$                & = & 1000.0 MPa \\ 
$\bar{\bar{\sig}}_f$ & = & 2.35 MPa \\ 
$\bar{\bar{\beta}}$  & = & 23.5 MPa/mm \\ 
\end{tabular} 
\caption{Parameter's values for reference simulation.} 
\label{tab_optimum} 
\end{table}

\subsection{Identification of elastic parameters} 
 
\begin{figure}[h] 
\center 
\includegraphics*[width=10cm,keepaspectratio]{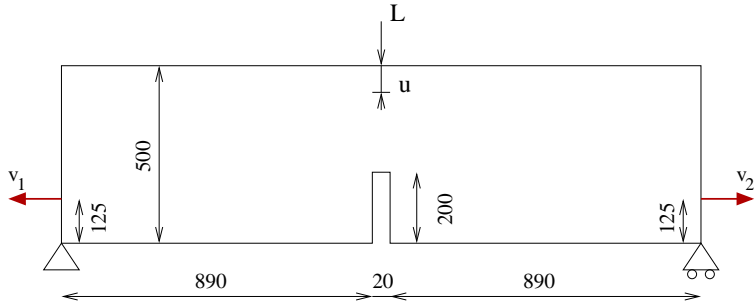} 
\caption{Displacements measured to evaluate the expansion $\Delta l = v_2 - v_1$ of the specimen.} 
\label{fig_bend-mesh1} 
\end{figure}

In the elastic range, Young's modulus and Poisson's ratio are
determined using a short simulations describing only the elastic
response of a specimen. Similarly to the uniaxial tensile test, the
elastic behavior is represented by the linear part of load-deflection
diagram. To identify both elastic parameters this information needs to
be supplemented with an additional measurement. In particular, we
propose to include the specimen expansion $\Delta l$ defined as the
relative horizontal displacements between the left and the right edge
of~the~specimen (as indicated by arrows in the Figure
\ref{fig_bend-mesh1}), or in other words $\Delta l = v_2 - v_1$. The
reference expansion-deflection curve is shown in Figure
\ref{fig_bend-ref}b. 
 
The objective function $F_1$ applicable for the determination of elastic parameters can be 
defined as follows: 
\begin{equation} 
F_1 = (L_{ref}(u)-L(u))^2 w_1 + (\Delta l_{ref}(u)-\Delta l(u))^2 w_2 \quad ; \qquad u=0.01 \mbox{mm} 
\label{eq_of1} 
\end{equation} 
The load $L$ and the expansion $\Delta l$ are quantities depending not
only on displacement $u$, but also on the values of material
parameters. In particular, at the beginning of the loading regime
(where $u = 0.01 \mbox{mm})$, the important parameters are only
Young's modulus and~Poisson's ration, because the other parameters are
not yet activated.  The quantities with index $_{ref}$ correspond to
the values taken from the reference diagram. The corresponding value
of weights $w_1$ and $w_2$ were calculated using 30 random simulations
to~normalize the average value of each of summation terms in
(\ref{eq_of1}) to one. It is worth noting that all the~quantities in
the objective function are evaluated for the prescribed deflection
$u=0.01$ mm, which allows the simulation to be stopped after reaching
this value. Therefore, the~first optimization stage is computationally
very efficient.
 
\begin{figure}[h] 
\center 
\begin{tabular}{cc} 
\includegraphics*[width=7cm,keepaspectratio]{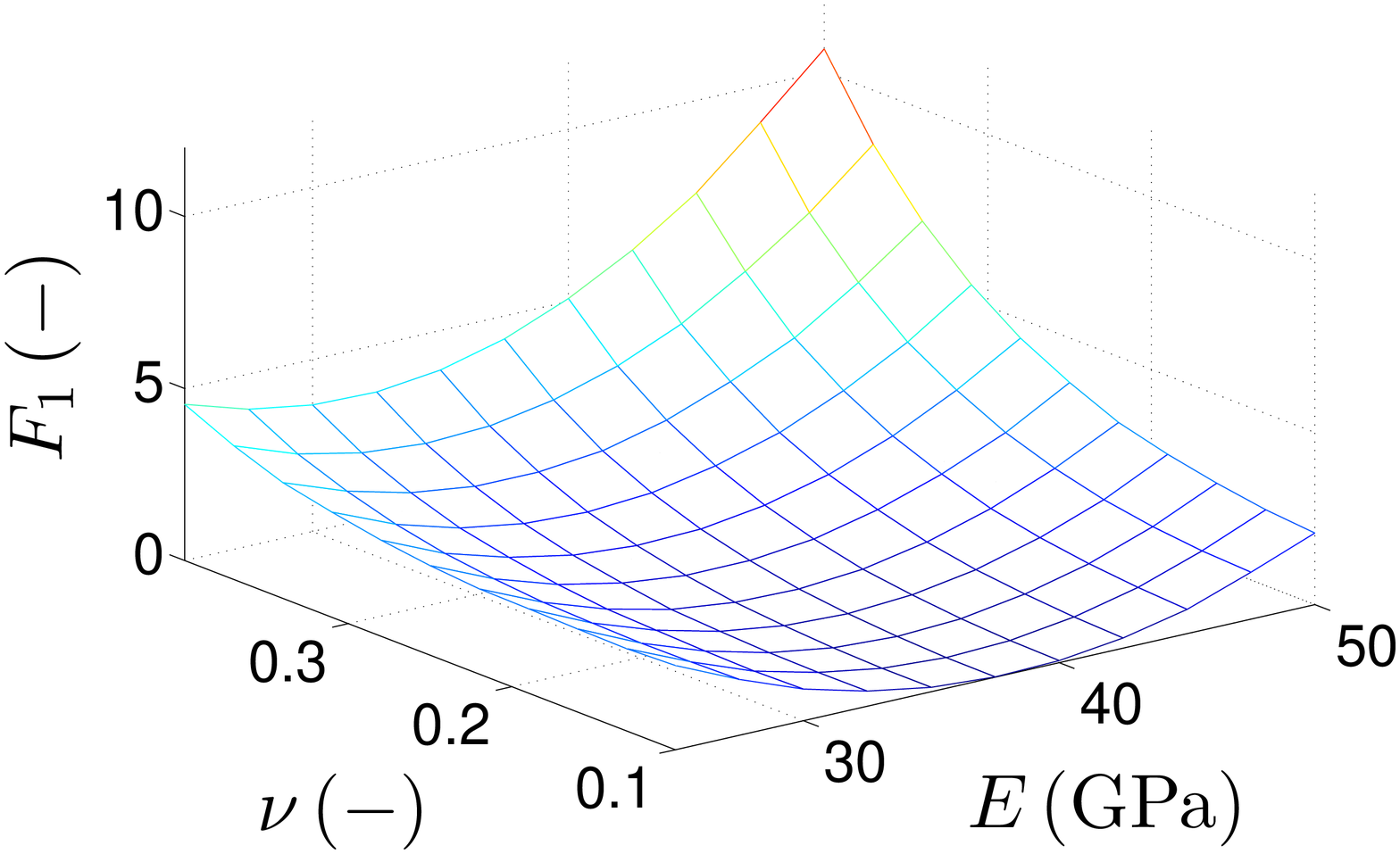} & 
\includegraphics*[width=7cm,keepaspectratio]{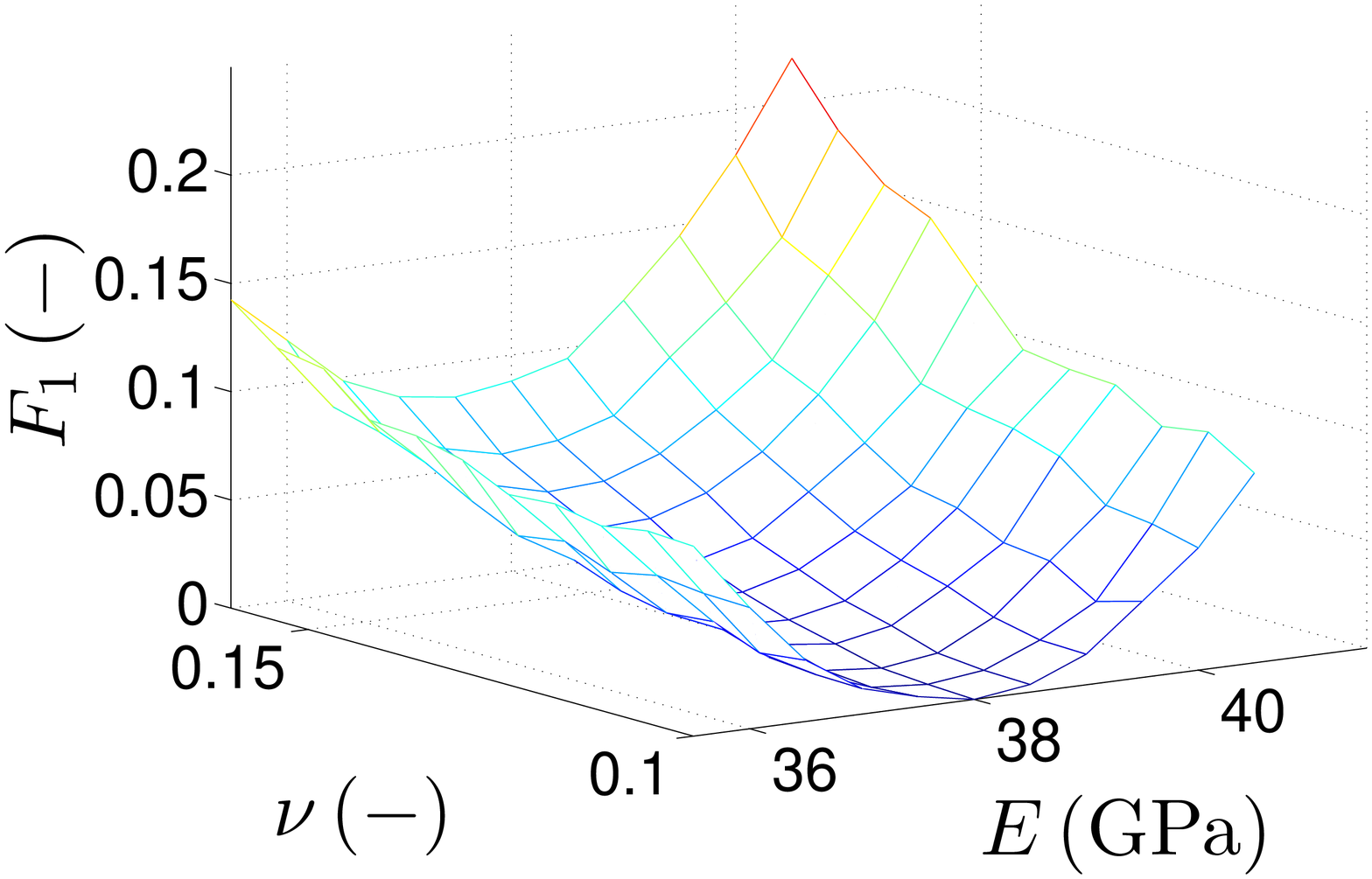} \\ 
(a) & (b) 
\end{tabular} 
\caption{Objective function $F_1$: (a) Whole domain (b) Detail 
close to optimal value.} \label{fig_of1} 
\end{figure}

For the sake of illustration, the shape of objective function 
$F_1$ is shown in the Figures \ref{fig_of1}a and \ref{fig_of1}b. 
As shown in this figure, the objective function remains rather 
wiggly in~the~neighborhood of the optimal value. 
 
\subsection{Identification of hardening parameters}\label{hard} 
 
\begin{figure}[h] 
\center 
\includegraphics*[width=10cm,keepaspectratio]{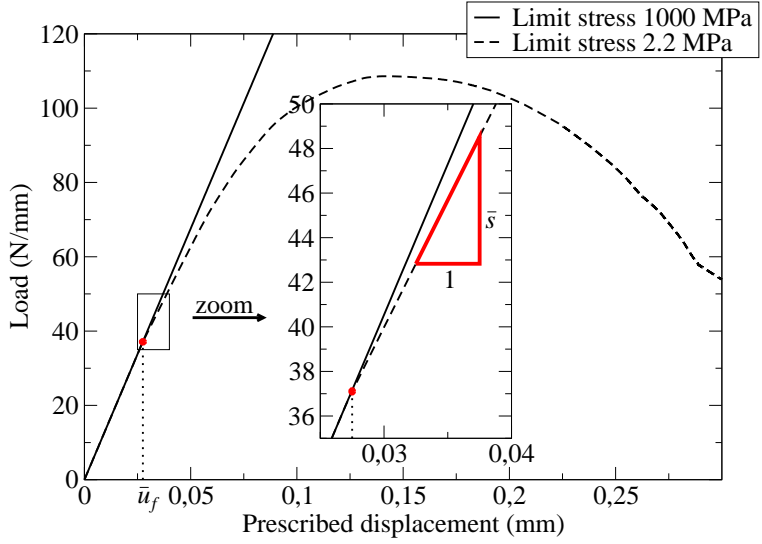}  
\caption{Measurements for objective function $F_2$ definition.} 
\label{fig_schema3} 
\end{figure} 

Once we have successfully determined Young's modulus and Poisson's
ratio, we can continue towards the estimate of the elastic limit
stress $\bar{\sig}_f$ (representing a threshold of elastic behavior)
and the hardening parameter $\bK$. In the spirit of the uniaxial
tensile test the limit stress will be related to the limit
displacement $\bar{u}_f$ at the end of the linear part of the
load-deflection diagram.  The hardening parameter $\bK$ will then
govern the slope of the diagram in the hardening regime. In our
particular case, the slope $\bar{s}$ is approximated as a secant
determined from two distinct points, see Figure
\ref{fig_schema3}. There are two contradictory requirements for that
choice: first, the points should not be too close to $\bar{u}_f$ to
ensure that numerical errors due to pseudo-time discretization do not
exceed the impact of $\bK$ parameter; second, they should be close
enough to $\bar{u}_f$ to ensure that the specimen does not enter 
the~softening regime and $\bar{\bar{\sigma}}_f$ will not be reached. 
If the~second requirement is fulfilled the corresponding objective
function depends only on values of the elastic limit stress
$\bar{\sigma}_f$ and the hardening parameter $\bar{K}$, because
Young's modulus and Poisson's ratio are fixed on the optimal values
determined during the previous optimization stage. The particular
choice objectie function adopted in this work is
\begin{equation} 
\bar{s} = (L(\bar{u}_f + 0.01 \mbox{mm}) - L(\bar{u}_f + 0.005 \mbox{mm}))/ 0.005  \mbox{mm} 
\label{eq_slope} 
\end{equation} 
leading to the objective function in form 
\begin{equation} 
F_2 = \left(\bar{u}_{f,ref} - \bar{u}_f \right)^2 w_3 + \left( \bar{s}_{ref} - \bar{s} \right)^2 w_4. 
\label{eq_of2} 
\end{equation}

\begin{figure}[h] 
\center 
\begin{tabular}{cc} 
\includegraphics*[width=7cm,keepaspectratio]{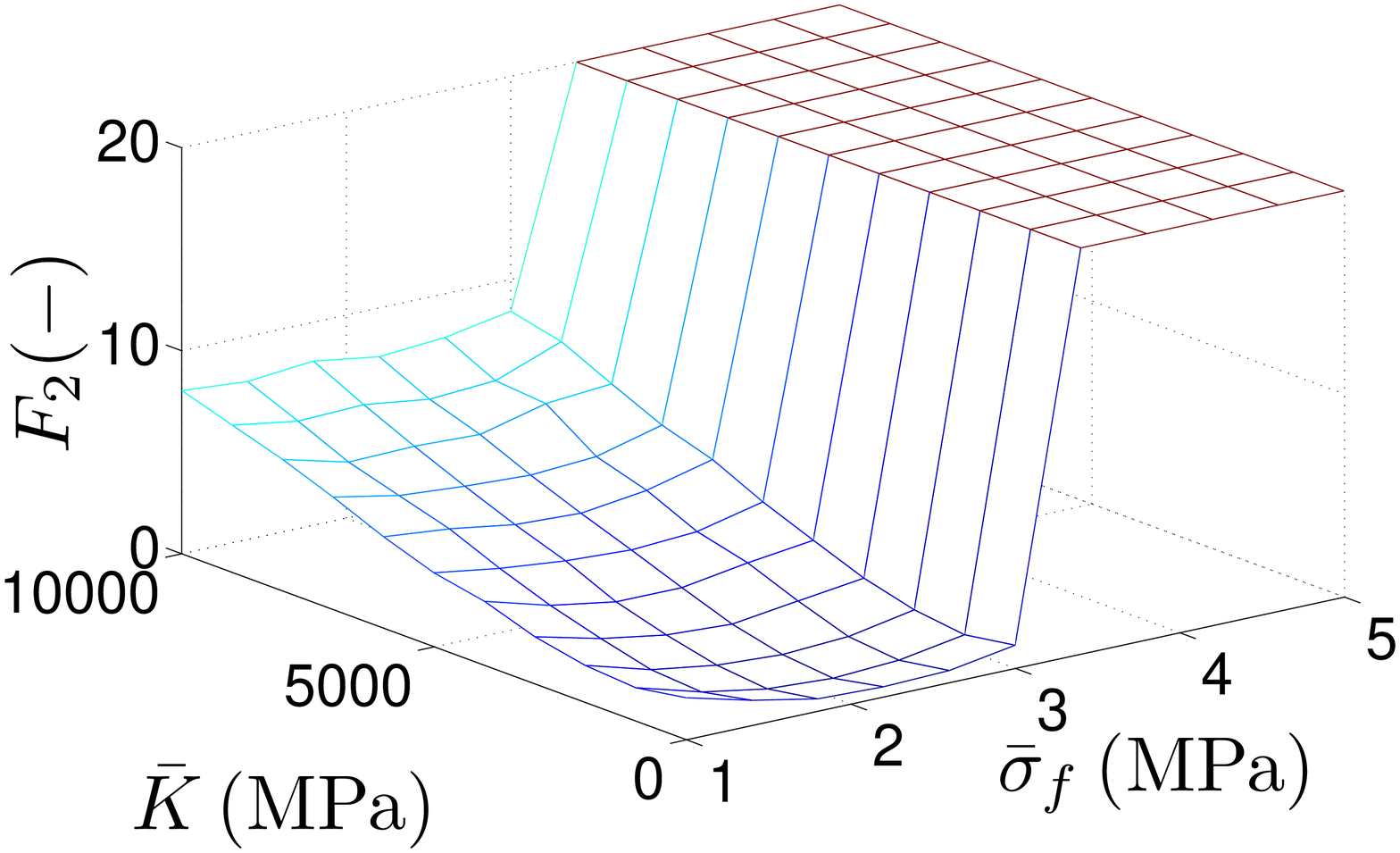} & 
\includegraphics*[width=7cm,keepaspectratio]{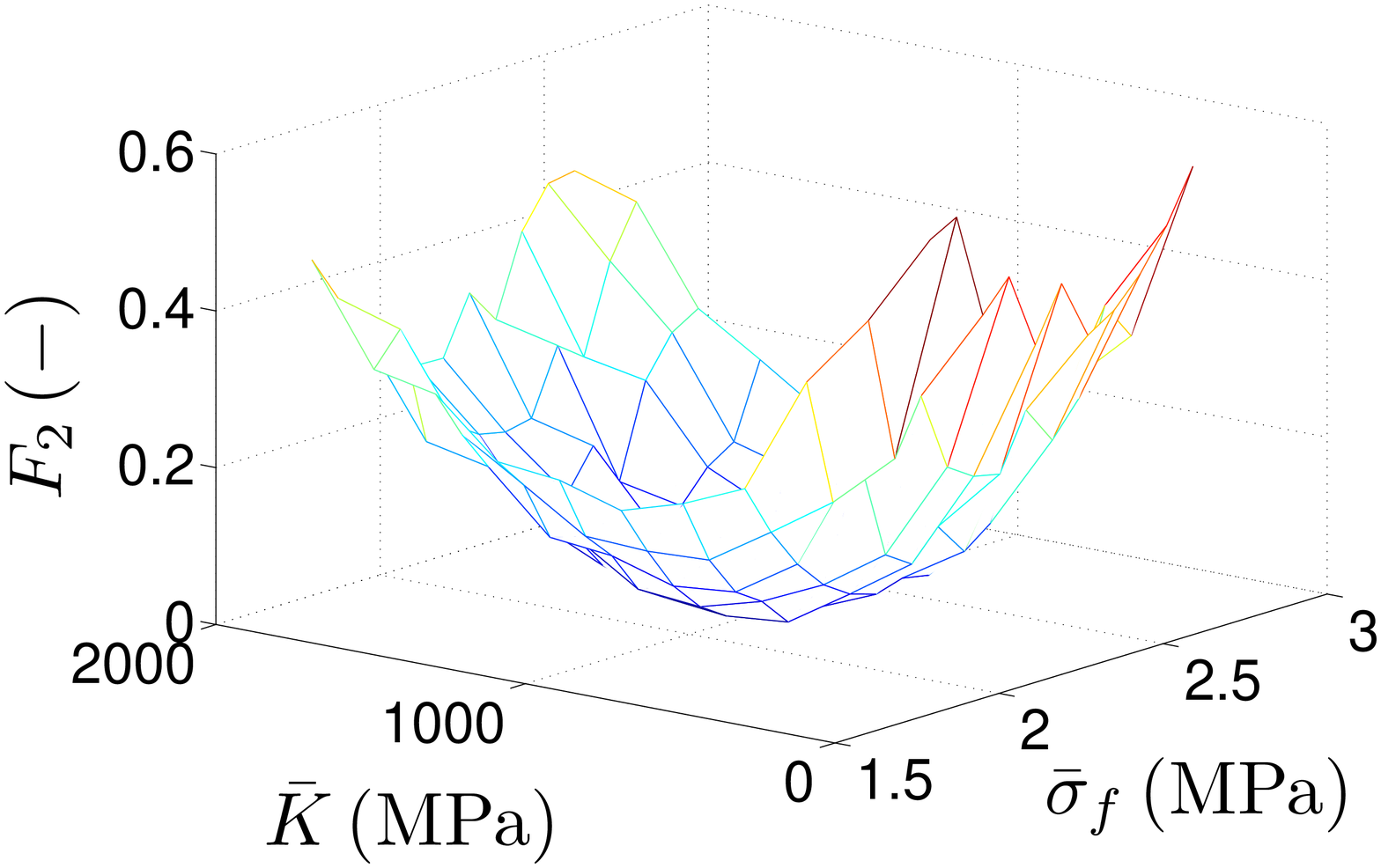} \\ 
(a) & (b) 
\end{tabular} 
\caption{Objective function $F_2$: (a) Whole domain (b) Detail close to optimum.} 
\label{fig_of2} 
\end{figure}

To keep this optimization step efficient, the simulations should 
again be restricted to~a~limited loading range, where the limit 
displacement can be related to the value of~$\bar{u}_{f,ref}$ from 
the reference diagram. Note that during optimization process, 
there is no guarantee that $\bar{\sig}_f$ will be exceeded when 
subjecting the specimen to the limit displacement. Such a~solution 
is penalized by setting the objective function value to $10 \times 
N$, where $N=2$ is the~problem dimension, see Figure 
\ref{fig_of2}a. Moreover, as documented by Figure \ref{fig_of2}b, 
the objective function is now substantially noisier than for the 
elastic case and hence more difficult to~optimize. 
 
\subsection{Identification of softening parameters}\label{soft} 
 
\begin{figure}[h] 
\center 
\includegraphics*[width=10cm,keepaspectratio]{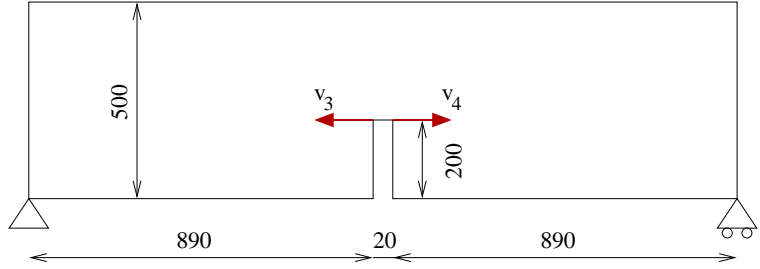} 
\caption{Displacement measured to express crack opening defined as $v_4 - v_3$.} 
\label{fig_bend-mesh2} 
\end{figure}

\begin{figure}[h]
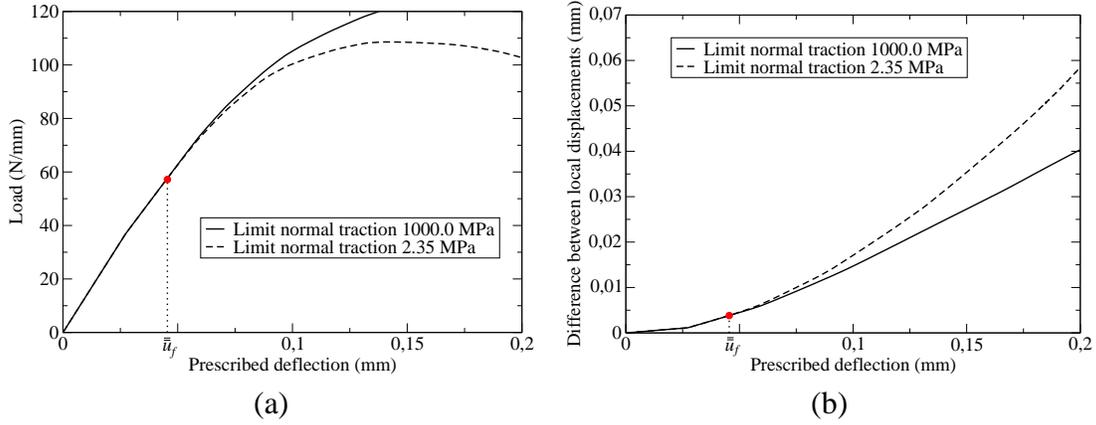
 
\center 
\begin{tabular}{cc} 
\includegraphics*[width=7cm,keepaspectratio]{bend/Stage3/fu-comp.eps} 
& 
\includegraphics*[width=7cm,keepaspectratio]{bend/Stage3/du-comp.eps} \\ 
(a) & (b) 
\end{tabular} 
\caption{Comparison of diagrams with and without the spring of 
crack (a) Load-deflection diagram (b) Evolution of difference 
between chosen local displacements during the loading.} 
\label{fig_s3-study} 
\end{figure}

The last stage of identification involves the discrete damage 
parameters: the limit normal traction $\bar{\bar{\sig}}_f$ and the 
softening parameter $\bar{\bar{\beta}}$. Variable 
$\bar{\bar{\sig}}_f$ represents a limit of hardening of~material 
and the appearance of a macroscopic crack. Determination of 
displacement~$\bar{\bar{u}}_f$ corresponding to this event, 
however, is rather delicate. The most straightforward method is 
based on the comparison of the reference curve with a simulation 
for a very high value of~$\bar{\bar{\sig}}_f$. The point where 
these two curves start to deviate then defines the wanted value 
of~$\bar{\bar{u}}_f$, see Figure \ref{fig_s3-study}a. A more reliable 
possibility could be based on a local optical measurement in the 
vicinity of notch \cite{Claire:2004} or acoustic emission 
techniques \cite{Chen:2004}. In our computations we consider local 
measurements of notch upper corners displacements $v_3$ and $v_4$, see Figure 
\ref{fig_bend-mesh2}. As demonstrated by graph 
\ref{fig_s3-study}b, the 'local' value of $\bar{\bar{u}}_f$ 
corresponds to the 'global' quantity rather well. 
 
After the specimen enters the softening regime, the shape of both 
local and global curves is fully governed by the softening parameter 
$\bar{\bar{\beta}}$. Therefore, its value can be fitted on~the~basis 
of the load corresponding to the deflection for which the softening is 
sufficiently active. This leads to the last objective function in the 
form 
\begin{equation} 
F_3 = \left( \bar{\bar{u}}_{f,ref} - \bar{\bar{u}}_f \right)^2 w_5 + \left( L_{ref}(u) - L(u) \right)^2 w_6 \quad ; \qquad u = 0.15 mm. 
\label{eq_of3} 
\end{equation} 
Since all the other parameters are already fixed on the values
determined during the previous optimization steps, this objective
function depends again only on two parameters: the limit normal
traciton $\bar{\bar{\sigma}}_f$ and the softening parameter
$\bar{\bar{\beta}}$.
 
\begin{figure}[h] 
\center 
\begin{tabular}{cc} 
\includegraphics*[width=7cm,keepaspectratio]{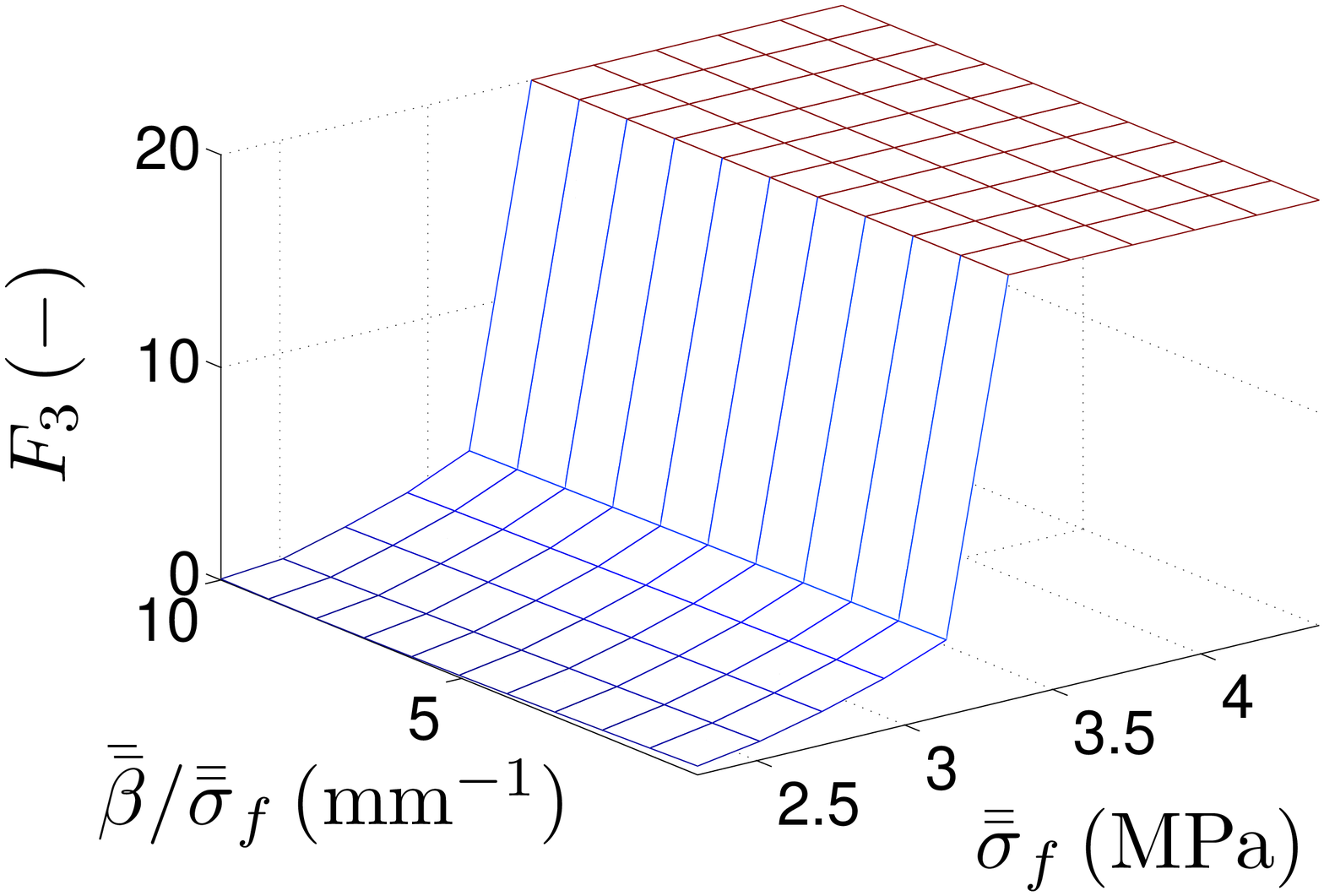} & 
\includegraphics*[width=7cm,keepaspectratio]{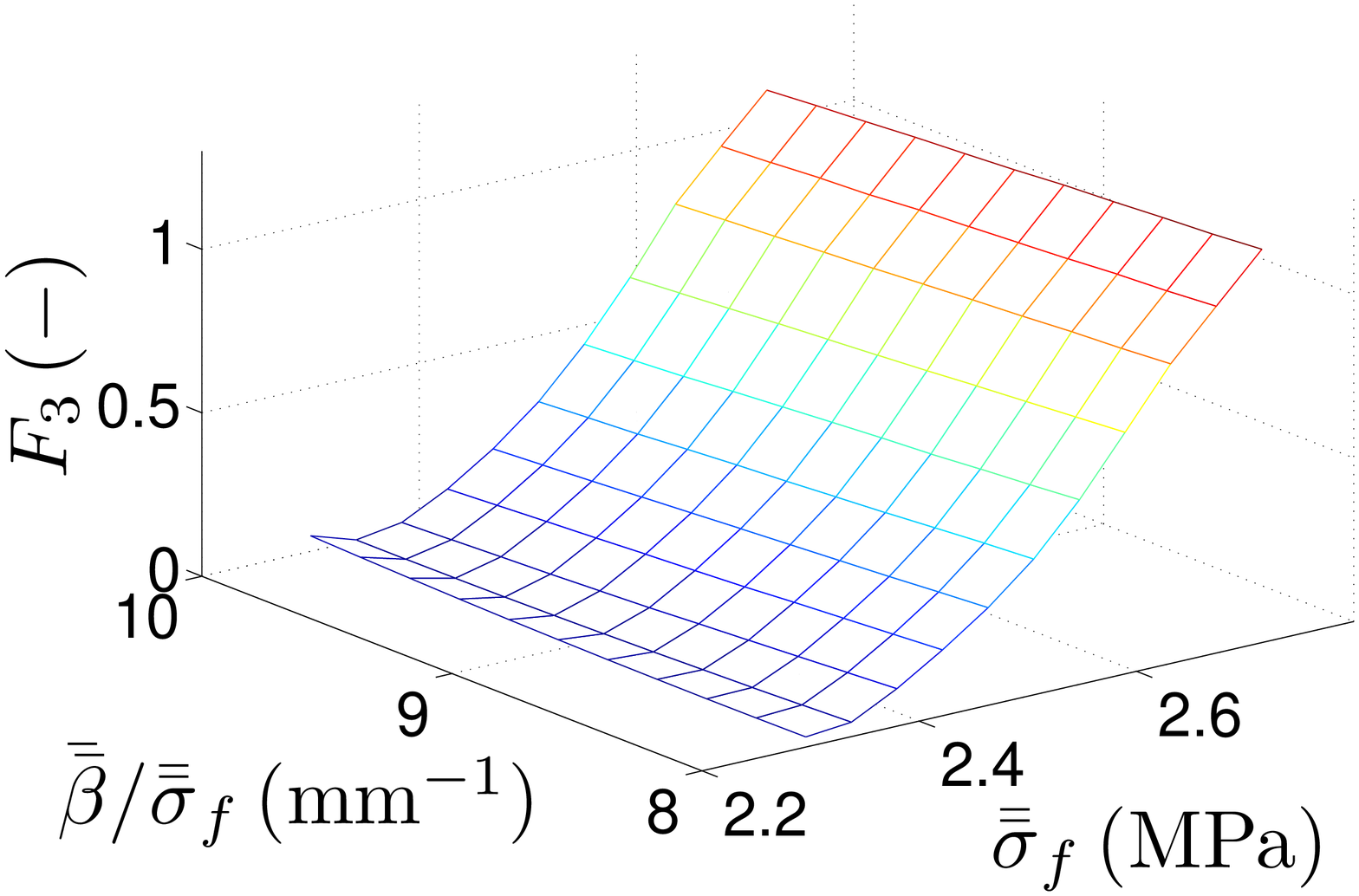} \\ 
(a) & (b) 
\end{tabular} 
\caption{Objective function $F_3$: (a) Whole domain (b) Detail 
close to optimal value.} \label{fig_of3} 
\end{figure}

By analogy to the procedure described in the previous section, the 
penalization is applied to the cases where the specimen does not 
reach the softening regime until the end of~simulations. This 
effect is well-visible from the surface of the objective function 
shown in~Figure~\ref{fig_of3}. 
 
\section{Optimization method}\label{optim} 
 
The algorithm used for the optimization of objective functions $F_1$, 
$F_2$ and $F_3$ is based on~an~efficient combination of an artificial 
neural network, namely the radial basis function network (RBFN), and 
an evolutionary algorithm {\bf GRADE} extended by the niching strategy 
CERAF. The principle of the algorithm is the replacement of an 
objective function by a neural network approximation and its 
subsequent optimization by an evolutionary algorithm. The approximation 
is constructed on a basis of interpolation of several points, where 
the values of objective function were calculated exactly. The 
approximation is adaptively improved by new neurons (points), 
 provided e.g. by optima located on~the~previous approximation. 
 
The main advantage of this methodology is an inexpensive 
evaluation of the approximation, which is repeatedly used during 
stochastic optimization process. Computationally expensive objective 
function is evaluated only when new neurons are added to the neural 
network. 
 
\subsection{Evolutionary algorithm GRADE extended by strategy CERAF}\label{GRADE} 
Evolutionary algorithms nowadays belong to the most popular 
optimization tools. Unlike the traditional gradient optimization 
methods, evolutionary algorithms operate on~a~population (a set of 
possible solutions), applying 'genetic' operators (cross-over, 
mutation and selection). The principles of evolutionary algorithm 
were first proposed by Holland~\cite{Holland:1975:ANS}. Ever 
since, the evolutionary algorithms have reached wide application 
domain, see e.g.~books by Goldberg~\cite{Goldberg:1989:GA} and 
Michalewicz~\cite{Michalewicz:1999:GA} for an extensive review. In 
this work, we employ a {\bf GRADE} algorithm (GRadient-based 
Atavistic Differential Evolution) developed in 
\cite{Ibrahimbegovic:2004}. The algorithmic scheme of the method 
is briefly summarized below. 
 
The first step is to~generate a starting generation of chromosomes by 
choosing random values of all state variables. The size of the 
population is in our case set to ten times the~number of optimized 
variables.  After the initiation stage the following steps are 
repeated until a termination condition is reached: 
 
\begin{enumerate} 
\item {\tt MUTATION}, which creates a 
new solution $\vek{x}$ using the operation 
\begin{equation} 
\vek{x}=\vek{y}+k(\vek{y}-\vek{z})\ , 
\end{equation} 
where $\vek{y}$ is a solution from actual population, $\vek{z}$ is a 
randomly created solution and~$k$ is a real number from given bounds. 
Ten percents of solutions in a new population are created by this operator. 
 
\item {\tt CROSS-OVER}, which creates a~new 
solution $\vek{x}$ according to 
\begin{equation} 
\vek{x}=\mbox{opt}(\vek{y},\vek{z})+cs(\vek{y}-\vek{z})\ , 
\label{eq_cross} 
\end{equation} 
where opt$(\vek{y},\vek{z})$ denotes a better solution from two 
randomly chosen vectors $\vek{y}$ and~$\vek{z}$, $c$ is a real random 
number with a~uniform distribution on~the~interval [0;2] and~$s$ 
changes the~direction of~the~descent $(\vek{y}-\vek{z})$ favor 
better solution from vectors $\vek{y}$ and~$\vek{z}$. This operator 
creates $90\%$ of individuals in a new population. 
 
\item {\tt EVALUATION} computes the objective 
function value for each new solution. 
 
\item {\tt TOURNAMENT SELECTION}, where the worst individual from two randomly 
selected solutions is deleted. This operator is repeated until the 
number of solution is the same as at the beginning of the cycle. 
 
\end{enumerate} 
 
The basic version of {\bf GRADE} is complemented with {\bf CERAF} 
strategy \cite{Hrstka:2004:AES} in order to increase the algorithm 
robustness when dealing with multi-modal problems.  The {\bf 
CERAF} strategy works on the principle of multi-start. If the best 
value found by a stochastic algorithm does not change for more 
than a prescribed value during a certain number of~generations, 
{\bf CERAF} store the optimal point in memory and surrounds it by 
a 'radioactive'' zone defined as $n$-dimensional ellipsoid with 
each diameter equal to 50\% of the size of~searched domain. If a 
new solution is located inside the 'radioactive'' zone, it is 
replaced with a random one outside the zone. If this solution was 
created by the {\tt CROSS-OVER} operator, the~size of the 
'radioactive'' zone is decreased by 0.5\% of the actual size. 
During the computations presented herein, the new local extreme 
was marked when the best value found had not changed for more than 
the stopping precision (defined as a property of objective 
function) during 100 generations.  An interested reader is 
referred to \cite{Hrstka:2004:AES} for~more detailed description 
of {\bf CERAF}. 
 
\subsection{Radial Basis Function Network}\label{rbfn} 
 
Artificial neural networks~(ANNs) were initially developed to 
simulate the processes in~a~human brain and later generalized for 
many problems like pattern recognition, different approximations 
and predictions, control of systems, etc 
\cite{Yagawa:1996:NNCM,Waszczyszyn:2006:NIA}. In this work, they 
will be used 'only'' as general approximation tools. The 
particular implementation is based on~the~idea of radial basis 
function networks as proposed e.g. in 
\cite{nakayama04,karakasis04}. 
 
\begin{figure}[!ht] 
\centering 
\scalebox{0.6}{ 
\includegraphics{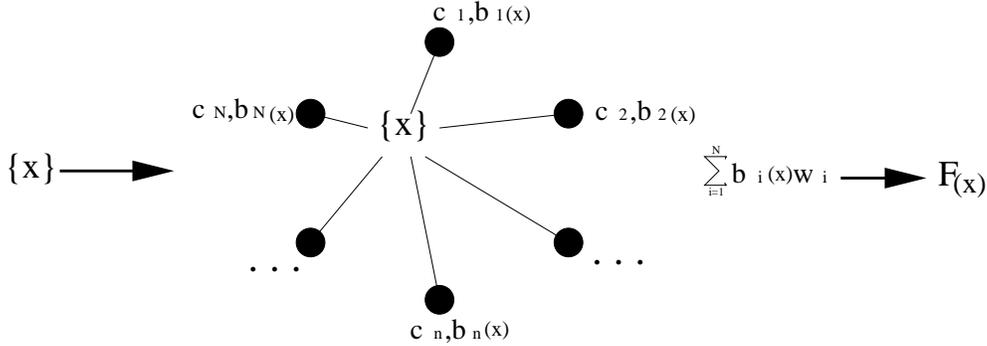}} 
\caption{An approximation using RBFN} 
\label{f:approx} 
\end{figure}

Neural network replace the objective function $F({\bf x})$ by approximation 
$\tilde{F}({\bf x})$ defined as~a~sum of radial basis functions multiplied 
by synaptic weights, see Figure \ref{f:approx}, 
 
or 
\begin{equation} 
F(\vek{x}) \approx \tilde{F}(\vek{x})=\sum_{i=1}^N b_i(\vek{x}) w_i\ , 
\label{approx} 
\end{equation} 
where $\vek{x}$ is a vector of unknowns, $b_i(\vek{x})$ is a~basis 
function associated with $i$-th neuron, $w_i$ is a~weight of the 
$i$-th neuron and $N$ is the total number of neurons creating the network. 
The basis function $b_i$ has the following form 
\begin{equation} 
b_i(\vek{x})=e^{-\parallel \vek{x}-\vek{c}_i\parallel^2/r}\ , 
\end{equation} 
where $\vek{c}_i$ is a vector  of the center coordinates for the 
$i$-th basis function and $r$ is a normalizing factor set to 
\begin{equation} 
r=\frac{d_{max}}{\sqrt[D]{DN}}\ , 
\label{norma} 
\end{equation} 
where $d_{max}$ is the maximal distance within the domain and $D$ is the 
number of dimensions. 
 
Synaptic weights are computed from the condition 
\begin{equation} 
F(\vek{c}_i)=\tilde{F}(\vek{c}_i), 
\end{equation} 
imposing the equality of the approximation and objective function 
values $\overline{y_i}$ in all neurons. This leads to a minimization 
problem in the form: 
\begin{equation} 
\min \sum_{i=1}^N[(\overline{y_i}-\sum_{j=1}^N b_j(\vek{c}_i) 
w_j)^2+\lambda_i w^2_i]\ . 
\label{MinE2} 
\end{equation} 
where $\lambda_i$ is a regularization factor set to $10^{-7}$. The 
solution of  (\ref{MinE2}) leads to a system of~linear equations 
determining the values of synaptic weights $\vek{w}$. 
 
At this point, the RBFN approximation of the objective function is 
created and the above-mentioned evolutionary algorithm GRADE is 
used to locate the 'approximate' global optima. In the next step, 
to improve the quality of approximation, three new points are 
added into the neural network: the optimum of previous 
approximation, a random point and another point in the descent 
direction defined by optima found in two previous steps. A more 
detailed description of the method is available in 
\cite{Kucerova:2005:topping}. 
 
In the current contest, one additional modification is introduced to 
deal with the penalization introduced in Sections \ref{hard} and 
\ref{soft}. If the point to be added to the neural network is 
penalized, it is accepted only in the case, when it was the optimum 
found by evolutionary algorithm. The rationale behind this 
modification is to avoid the penalty-imposed discontinuities in the 
objective function, which are impossible to capture by the smooth RBFN 
approximation.

\section{Identification procedure verification}\label{results} 
 
When assessing the reliability of the proposed identification procedure 
a special care must be given to stochastic nature of the optimization 
algorithm. The analysis presented herein is based on the statistics of 
100 independent optimization processes, executed for each objective 
function. The termination criteria were set to: 
\begin{itemize} 
\item the number of objective function evaluations exceeded 155; 
\item the value of objective function smaller than a stopping precision was found. 
\end{itemize} 
A particular optimization process was marked as 'successful', when 
the latter termination condition was met.  Note that since the 
reference simulation instead of experimental data are used, the 
optimum for every objective function is equal to zero. Results of 
the performance study are summarized in Table \ref{tab_precval} 
showing the success rate related to a stopping precision together 
with the maximum and average number of function evaluations. 
\begin{table} 
\begin{tabular}{ccccc} 
$F$   & Stopping precision & Successful runs & Maximal number      & Average number        \\ 
       &                    &                  & of $F$'s evaluation  & of $F$'s evaluation    \\ 
\hline 
$F_1$ & $10^{-3}$ & 100 & 32 & 16 \\ 
\hline 
$F_2$ & $10^{-2}$ &  94 & 140 & 29 \\ 
$F_2$ & $10^{-3}$ &  80 & 140 & 47 \\ 
\hline 
$F_3$ & $10^{-2}$ &  92 & 140 & 37 \\ 
$F_3$ & $3.10^{-3}$ &  76 & 143 & 47 \\ 
\hline 
\end{tabular} 
\caption{Summary of reliability study.} 
\label{tab_precval} 
\end{table} 
 
Moreover, the different values of stopping precision allow us to investigate the 
relation between the accuracy of identified parameters and the 
tolerance of the objective function value. Table \ref{tab_ofpar} 
shows a concrete outcome of such an analysis, where the maximal 
and~average errors are calculated relatively to the size of the interval 
given by limit values for~each parameter. 
\begin{table} 
\begin{tabular}{ccccc} 
\hline 
Parameter & Stopping precision on $F$ & Average error [\%] & Maximal error [\%] \\ 
\hline 
$E$             & $10^{-5}$     & 0.41 & 1.23 \\ 
$\nu$           & $10^{-5}$     & 0.16 & 2.20 \\ 
\hline 
$\bar{\sigma}_f$    & $10^{-2}$     & 0.87 & 2.58 \\ 
$\bK$           & $10^{-2}$     & 0.78 & 2.49 \\ 
$\bar{\sigma}_f$    & $10^{-3}$     & 0.30 & 0.59 \\ 
$\bK$           & $10^{-3}$     & 0.49 & 1.54 \\ 
\hline 
$\bar{\bar{\sigma}}_f$  & $10^{-2}$     & 0.47 & 1.32 \\ 
$\bar{\bar{\beta}}$ & $10^{-2}$ & 2.34 & 12.21 \\ 
$\bar{\bar{\sigma}}_f$  & $3\times10^{-3}$  & 0.33 & 0.67 \\ 
$\bar{\bar{\beta}}$ & $3\times10^{-3}$  & 0.26 & 2.68 \\ 
\hline 
\end{tabular} 
\caption{Influence of stopping precision on accuracy of identified parameters.} 
\label{tab_ofpar} 
\end{table} 
 
\begin{figure}[h]
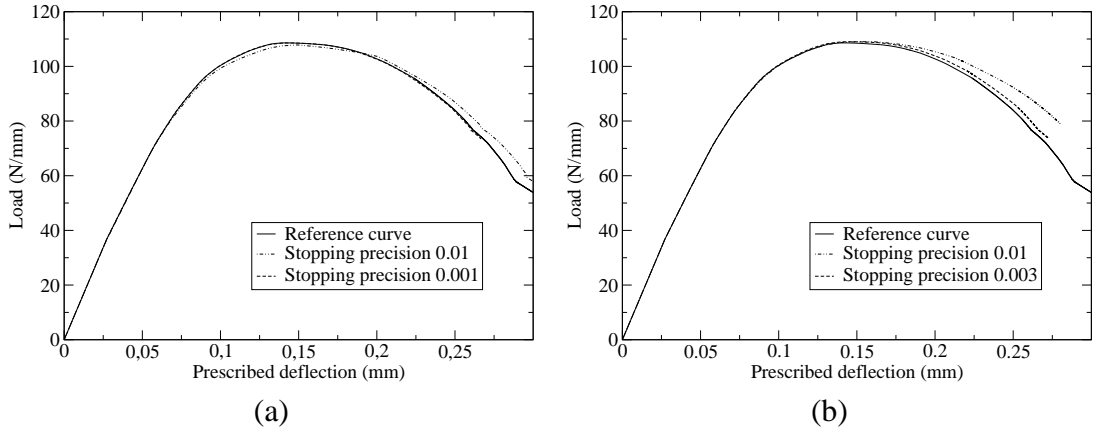
 
\center 
\begin{tabular}{cc} 
\includegraphics*[width=7cm,keepaspectratio]{bend/Stage2/fu-prec-comp.eps} & 
\includegraphics*[width=7cm,keepaspectratio]{bend/Stage3/fu-prec-comp.eps} \\
(a) & (b) 
\end{tabular} 
\caption{Comparison of load-deflection diagrams: (a) Hardening parameters (b) Softening parameters.} 
\label{fig_comp} 
\end{figure}

The results show that the maximal error for the elastic parameters $E$ and $\nu$ is less 
than $3\%$, which is sufficient from the practical point of view. This 
is also documented by Figure \ref{fig_comp}, where no deviation of the 
reference and resulting curves is visible in the elastic range. For the 
hardening stage (parameters $\bar{\sigma}_f$ and $\bK$) a similar 
precision is unfortunately not sufficient as shown by Figure 
\ref{fig_comp}a. 
Increasing the stopping precision to $10^{-3}$ reduces the~error on 
parameters roughly by $50\%$, which is sufficient to achieve almost 
perfect fit of~the~reference curve. 
 
Finally, a similar conclusion holds for the parameters 
$\bar{\bar{\sigma}}_f$ and $\bar{\bar{\beta}}$ describing the 
softening part of the experiment. The stopping precision equal to 
$10^{-2}$ is too coarse to achieve sufficient precision on parameters 
and needs to be reduced to $3\times10^{-3}$. The effect of~increased 
accuracy is then well visible in Figure \ref{fig_comp}b. This step 
completes the verification of~the~algorithm. 
 
\begin{figure}[h] 
\center 
\begin{tabular}{cc} 
\includegraphics*[width=8cm,keepaspectratio]{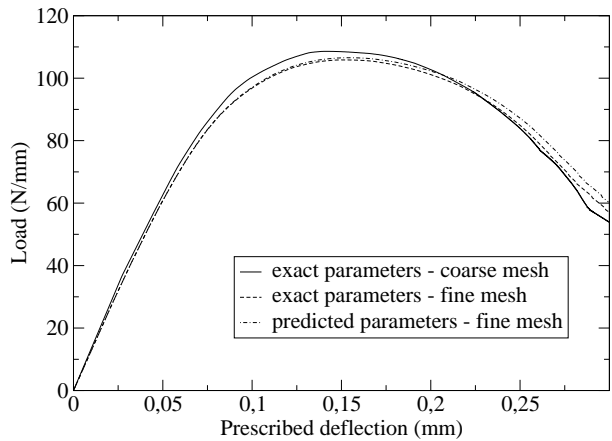} 
\end{tabular} 
\caption{Comparison of load-deflection diagrams:  computations 
with both coarse and fine mesh, for exact and predicted parameter 
values.} \label{fig_comp1} 
\end{figure}

One more analysis has been performed to estimate the importance of 
the mesh refinement for the material parameter identification. 
Namely, we have repeated the analysis on~the~refined mesh, by 
using both the exact parameter values and the estimates provided 
by~the~procedure proposed herein. The results are shown in Figure 
\ref{fig_comp1}, where we plot the~force-\-displacement diagram 
for coarse mesh and fine mesh based computations. We can see that 
the mesh refinement reduces the errors produced with the parameter 
estimation. The latter can be combined with the adaptive mesh 
procedure, which is currently being explored \cite{brancher2006}. 
 
\section{Conclusion}\label{conclu} 
 
We have proposed a sound identification procedure for material 
parameters of the constitutive model for representing the 
localized failure of massive structures. The most pertinent 
conclusions can be stated as follows: 
 
i) The sequential identification approach employed for the 
uniaxial tensile test can be extended to the three-point bending 
test. The resulting algorithm is very straightforward and has a 
clear link with the structure of the constitutive model. Moreover, 
each of three stages uses only a part of the test simulation, 
which leads to substantial computational time savings. 
 
ii) Due to the physical insight into the model, it was possible to 
construct simple objective functions $F_1$, $F_2$ and $F_3$ with a 
high sensitivity to the relevant parameters. This led 
to~non-smooth and non-convex objective functions, which were 
optimized by robust soft-computing methods. 
 
iii) The proposed identification procedure was verified on $100$ 
independent optimization processes executed for each objective 
function. In the worst case, the reliability of~the~algorithm is 
$76\%$ due to very small number of objective functions calls set 
in the termination condition. From our experience with 
evolutionary algorithms \cite{Hrstka:2003:CS}, such a result is 
rather satisfactory. 
 
iv) As the result of a sequential character of the identification 
procedure, the errors in~identified parameters accumulate. 
Therefore, the values need to be determined with higher accuracy 
then usually required in applications (i.e. $5\%$) and achievable 
by neural network-based inverse analysis 
\cite{Kucerova:2007:CAMES}. 
 
v) The major difficulty of the proposed methods is to properly 
identify the three stages of structural behavior. From the point 
of view of method verification, where the reference 
load-deflection diagram is not noisy, the problem was successfully 
resolved. To fully accept the procedure, however, the experimental 
validation of the method appears to be necessary. 
 
\section{Acknowledgments} 
The financial supports for this work by the Czech Ministry of 
Education (research project MSM~6840770003) and the French 
Ministry of Research are gratefully acknowledged. JZ acknowledges 
the support of the ENS-Cachan invited professor funding program.

\end{document}